\documentclass[10pt,twocolumn,letterpaper]{article}

\usepackage{wacv}
\usepackage{times}
\usepackage{epsfig}
\usepackage{graphicx}
\usepackage{amsmath, colortbl}
\usepackage{amssymb}
\usepackage{booktabs}
\usepackage{microtype}
\usepackage{enumitem}
\usepackage{appendix}

\newcolumntype{H}{>{\setbox0=\hbox\bgroup}c<{\egroup}@{}}

\def\wacvPaperID{670} 

\wacvalgorithmstrack   

\wacvfinalcopy 

\ifwacvfinal
\usepackage[breaklinks=true,bookmarks=false]{hyperref}
\else
\usepackage[pagebackref=true,breaklinks=true,colorlinks,bookmarks=false]{hyperref}
\fi

\begin{document}


\title{PointInverter: Point Cloud Reconstruction and Editing via a Generative Model with Shape Priors}

\author{Jaeyeon Kim $^1$ \qquad Binh-Son Hua $^2$ \qquad Duc Thanh Nguyen $^3$\qquad Sai-Kit Yeung $^1$\\\\
$^1$ Hong Kong University of Science and Technology \qquad $^2$ VinAI Research \qquad $^3$ Deakin University
}

\maketitle
\thispagestyle{empty}

\begin{abstract}
   In this paper, we propose a new method for mapping a 3D point cloud to the latent space of a 3D generative adversarial network. Our generative model for 3D point clouds is based on SP-GAN, a state-of-the-art sphere-guided 3D point cloud generator. We derive an efficient way to encode an input 3D point cloud to the latent space of the SP-GAN. Our point cloud encoder can resolve the point ordering issue during inversion, and thus can determine the correspondences between points in the generated 3D point cloud and those in the canonical sphere used by the generator. We show that our method outperforms previous GAN inversion methods for 3D point clouds, achieving state-of-the-art results both quantitatively and qualitatively. Our code is available at \url{https://github.com/hkust-vgd/point_inverter}.
\end{abstract}

\section{Introduction}
\label{sec:intro} 

Deep learning for 3D point clouds has been rapidly progressing in the past five years. A majority of research have been dedicated to design efficient ways for neural networks to process 3D point clouds~\cite{qi2017pointnet,qi2017pointnet++,li2018pointcnn,wang2019dgcnn,zhang-shellnet-iccv19}. These developments have empowered the learning ability of neural networks for a wide range of downstream tasks such as object classification, semantic segmentation, object detection on both synthetic and real-world data~\cite{wu20153d,chang2015shapenet,hua2016scenenn,dai2017scannet,uy-scanobjectnn-iccv19}.

Compared to point cloud analysis, research on generative modeling of 3D point clouds is more scarce. 
A notable work, namely SP-GAN~\cite{li2021spgan} demonstrates remarkable results in generating 3D point clouds from a shape prior using generative adversarial neural networks. Despite this important outcome, the generative model in the SP-GAN is unconditional, meaning that it only allows sampling of novel point clouds while leaving manipulating and editing of existing point clouds unexplored. 

A common approach to perform data editing with a generative adversarial network is to follow a two-stage principle: \emph{invert first, edit later}. This principle has been well demonstrated in the 2D domain, where state-of-the-art GAN inversion methods~\cite{richardson2021encoding,alaluf2021restyle,alaluf2021hyperstyle,dinh2022hyperinverter} are used to map a real image into the latent space of the StyleGAN model~\cite{karras2019stylegan2} by using optimization-based or encoder-based techniques. Once the mapping is done, the reconstructed latent code can be manipulated to make effects on the real image (e.g., changing attributes of the image). Motivated by this principle, we develop a new approach for inverting a 3D point cloud so that the point cloud's latent code can be faithfully used to reconstruct a new point cloud via the SP-GAN model. 

A key challenge in 3D deep learning for point cloud data is to design a learning mechanism that is invariant to point ordering. In point cloud analysis, this capability can be achieved by aggregating point features using a symmetric function~\cite{qi2017pointnet}, sorting the points using a grid before performing convolution operations~\cite{hua2017point}. However, point ordering has been largely ignored from point-cloud GAN inversion research~\cite{zhang2021shapeinversion}. The inversion is achieved by comparing reconstructed point clouds with their original point clouds using metrics such as Chamfer discrepancy or earth-mover distance, which is invariant to point ordering. 
However, this manner makes the correspondences between the reconstructed and original point clouds to no longer be maintained, limiting the ability of downstream applications, e.g., point cloud matching and registration. 

We address this problem by resolving point orders during inversion. Our method can be summarised as follows. Given a pretrained SP-GAN model, we design a point-based encoder that takes a 3D point cloud as input, then extracts two style vectors from the input. These style vectors, together with the point cloud's features are fed into the pretrained SP-GAN. At the first stage, we reconstruct a global latent code via jointly training the encoder and refining the generator. At the second stage, we use the global latent code to resolve point ordering. We then extract local latent codes from the global latent code at the last stage. Our method achieves state-of-the-art performance in reconstructing 3D point clouds, both quantitatively and qualitatively. For applications, we show that latent codes can be applied to determine correspondences between input point clouds and reconstructed results, enabling point cloud manipulation and editing ability. 

In summary, our contributions are as follows:
\begin{itemize}[leftmargin=*]
    \item A point cloud encoder that maps a 3D point cloud to the latent space of the SP-GAN, a state-of-the-art sphere-guided point cloud generator;
    \item A strategy to resolve point ordering during inversion by leveraging global latent codes generated by the point cloud encoder, thus maintaining point correspondences in the reconstruction and enabling shape editing;
    \item A global to local latent code refinement technique that better preserves both geometric details and correspondences in the reconstruction;
    \item State-of-the-art results in point cloud GAN inversion, illustrated via point cloud reconstruction and editing. 
\end{itemize}

The remainder of this paper is organized as follows. We review the related work in Section~\ref{sec:related}. In Section~\ref{sec:background}, we present the SP-GAN~\cite{li2021spgan}, which is used as a decoder reconstructing points clouds in our proposed GAN inversion method. Our method is then described in Section~\ref{sec:method}. Applications and experimental results are reported in Section~\ref{sec:experiment}. Section~\ref{sec:conclusion} concludes our work and provides remarks.

\section{Related Work} 
\label{sec:related}

\begin{figure*}
\centering
\includegraphics[width=0.9\linewidth]{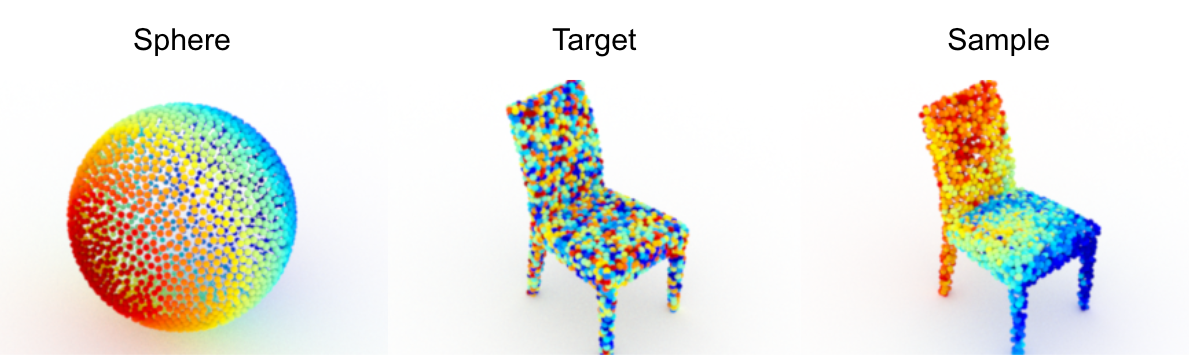}
\caption{An advantage of point cloud GAN inversion using a generator with a shape prior (SP-GAN~\cite{li2021spgan}) is that we can reconstruct both points and their correspondences to the shape prior. This figure illustrates the dense correspondence between the guided sphere, the target point cloud for inversion, and the reconstructed point cloud by our method. As can be seen, the point correspondences are random on the target point cloud before performing the inversion. Our reconstructed point cloud is geometrically similar to the target while having smooth correspondences.}
\label{fig:dense_correspondence}
\end{figure*}

\paragraph{3D deep learning.} Deep learning in the 3D domain has progressed tremendously in the past few years, with 3D point cloud deep learning as one of the main research directions due to the universality of the point cloud representation. Several point cloud neural networks are designed for analysis tasks such as object classification, semantic segmentation, and object detection. Notable works include PointNet~\cite{qi2017pointnet} as the pioneering method that first learns per-point features and then pools them into a global feature vector, PointNet++~\cite{qi2017pointnet++} and DGCNN~\cite{wang2019dgcnn} as popular methods to capture local point features, and a long list of other methods for building convolutions on point clouds~\cite{li2018pointcnn,xu2018spidercnn,zhang-shellnet-iccv19} and for handling downstream tasks such as semantic segmentation~\cite{huang2018recurrent,wang2018deep,graham20183d,landrieu2018large,hu2020randla,xu2020squeezesegv3}. In this work, we leverage DGCNN as a point cloud encoder in our method.

\paragraph{Generative modeling for 3D point clouds.}
Despite the rapid development of 3D deep learning~\cite{guo-point-survey-2019}, generative models for 3D point clouds are relatively scarce. To unconditionally generate a 3D point cloud, an early attempt is to train an autoencoder on 3D point clouds and then train a GAN in the latent space learned by the autoencoder~\cite{achlioptas2018latent}. Subsequent methods include the use of autoregressive model~\cite{sun2020pointgrow}, flow-based generation~\cite{yang2019pointflow,kim2020softflow,klokov2020discrete,cai2020learning}, and generative adversarial neural networks~\cite{ramasinghe2019spectral,hui2020progressive,li2021spgan} with tree-based structures~\cite{shu2019treegan,gal2021mrgan}. Among those GAN-based methods, recently, SP-GAN~\cite{li2021spgan} learns to generate a point cloud by transforming a prior geometry such as a canonical sphere to a 3D shape. There also exists a group of works~\cite{fan2017point,arshad2020progressive} that learn to predict point clouds from conditional data such as images. 
Inspired by the great success of generative adversarial neural networks, in this work, we focus on point clouds generated by unconditional GAN models and how these models can be leveraged to manipulate and edit 3D point cloud data by using GAN inversion pipeline.

\begin{figure*}[t]
\centering
\includegraphics[width=1.0\linewidth]{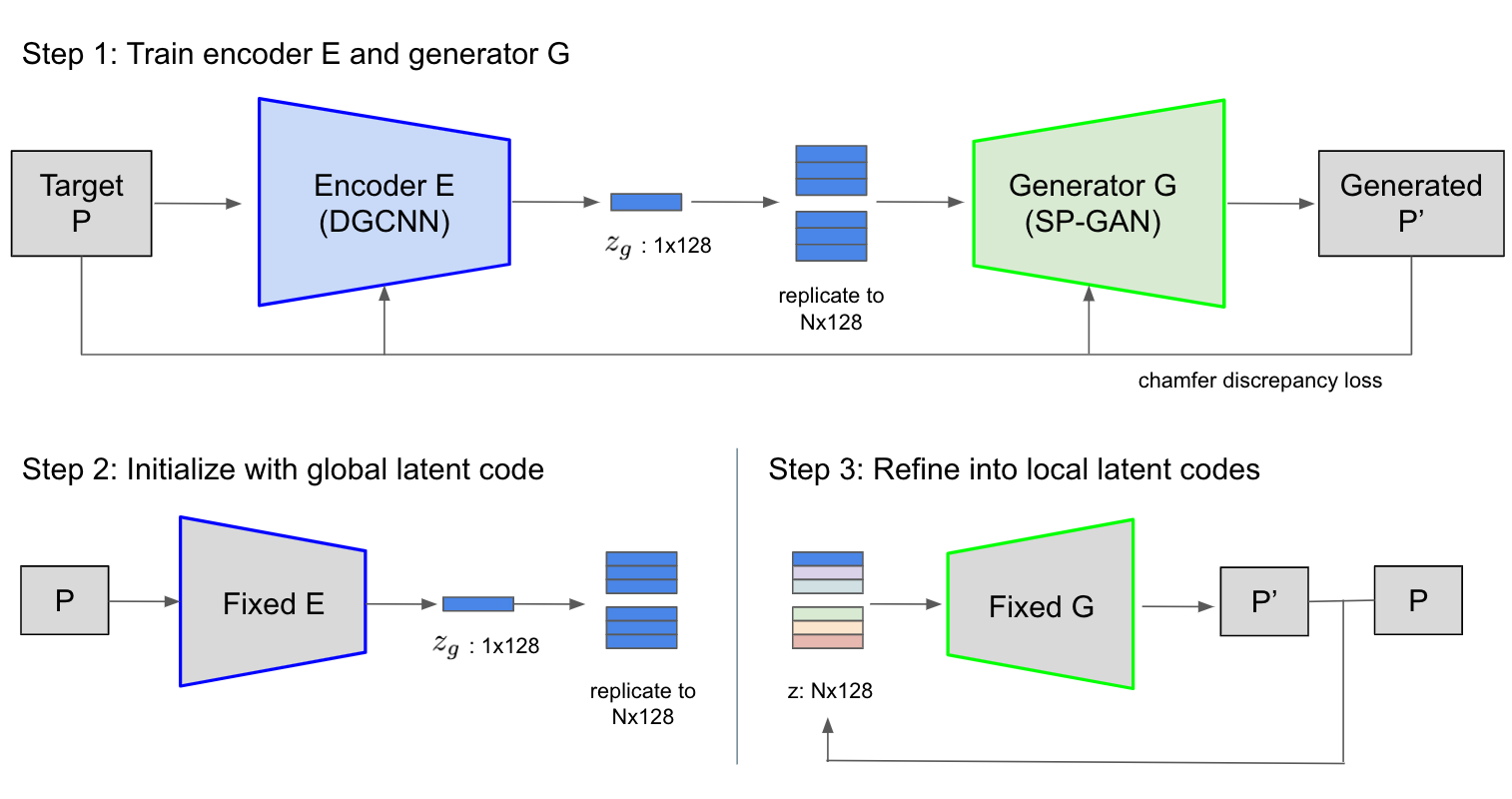}
\caption{Our GAN inversion for point clouds. The encoder is built upon the pre-trained DGCNN and the generator is based on the pre-trained SP-GAN. In Step 1, the encoder $E$ and generator $G$ are trained to map a global latent code $z_g$ to a target point cloud $P$. In Step 2, following SP-GAN, the global latent code $z_g$ is replicated by the number of points $N$ to initialize local latent codes. In Step 3, we refine the local latent codes by iteratively performing an optimization task.}
\label{fig:inversion_framework}
\end{figure*}

\paragraph{GAN inversion.}
The basic principle of GAN inversion is to faithfully map an input shape to the latent space of a GAN model. Manipulation and editing can then be done in the GAN latent space, depending on the downstream task. However, reconstructing the input using a pretrained GAN is challenging. In the image domain, StyleGAN~\cite{karras2019stylegan} has made a milestone. Typical inversion methods can be categorized as optimization-based methods or learning-based methods. Optimization-based methods include optimizing a simple projection function~\cite{karras2019stylegan} or fine-tuning a generator for each image~\cite{roich2021pivotal}. These methods provide high-quality reconstruction but also require heavy computation. In contrast to optimization-based methods, learning-based methods allow lighter and faster reconstruction by training an encoder to directly output the latent code~\cite{richardson2021encoding,alaluf2021restyle}, or by training a hypernetwork for defining a refined generator to improve reconstruction quality~\cite{alaluf2021hyperstyle,dinh2022hyperinverter}.

In the 3D domain, several attempts have been made for shape reconstruction using the latent space of a generative model for 3D point clouds. For example, Zhang et al.~\cite{zhang2021shapeinversion} proposed to use GAN inversion for 3D point clouds to address the shape completion task~\cite{chen2020pcl2pcl}. Their generative model was built upon the tree-GAN developed in~\cite{shu2019treegan}, where the generator was a graph convolutional network applied to a tree data structure. 
Our work differs from~\cite{zhang2021shapeinversion}. Specifically, we focus not only on the quality of the reconstruction but also point ordering. In our work, we adopt the SP-GAN developed in~\cite{li2021spgan} as our generative model. This generator leverages a geometric prior in the form of a canonical sphere to guide the shape generation process, which results in better shape quality and controllability.
\section{Background} 
\label{sec:background}

SP-GAN proposed in~\cite{li2021spgan} is a neural network architecture aiming to reconstruct point clouds $P$ from a spherical structure. Like standard unconditional GANs, the SP-GAN also includes an unconditional generator $G$ and a discriminator $D$. The generator $G$ takes input as a unit sphere and generates a point cloud $P$. The discriminator $D$ takes input as a point cloud either generated by the generator $G$ or sampled from training data, and classifies the input point cloud and all of its points into two classes: real vs fake. Both the generator $G$ and discriminator $D$ are trained simultaneously in an  alternating manner. We present the training and testing of the SP-GAN in detail below.

Let $S$ be a unit sphere including $N$ points; these points are evenly distributed on $S$. The spatial locations (i.e., 3D coordinates) of these $N$ points are used to define a global prior in $\mathbb{R}^{N\times 3}$. Each point is also associated with a local prior, which is a random noise vector in $\mathbb{R}^d \sim \mathcal{N}(0,1)$, following a standard normal distribution. The sphere $S$ is finally encoded into a prior latent code $z_S \in \mathbb{R}^{N\times (3 + d)}$ by concatenating 3D coordinates and local prior for every point in $S$.

The prior latent code $z_S$ is then used to train the generator $G$ for constructing point clouds $P$, each of which also contains $N$ points. While training the generator $G$, real point clouds $P'$ are sampled from a training dataset and used to train the discriminator $D$. 

The SP-GAN can be trained end-to-end by simultaneously minimizing a discriminative loss $\mathcal{L}_D$ and a generative loss $\mathcal{L}_G$. The discriminative loss $\mathcal{L}_D$ includes two sub-losses: $\mathcal{L}^{\text{point cloud}}_D$ for an entire point cloud and $\mathcal{L}^{\text{point}}_D$ for individual points. These losses are defined as follows. 
\begin{align}
\label{eq:SPGAN_Discriminative_Loss}
\mathcal{L}_D &= \mathcal{L}^{\text{point cloud}}_D + \lambda \mathcal{L}^{\text{point}}_D, \\
\mathcal{L}^{\text{point cloud}}_D &= \frac{1}{2} [(D(P))^2 + (D(P')-1)^2], \\
\mathcal{L}^{\text{point}}_D &= \frac{1}{2N} \sum_{i=1}^N [(D(p_i))^2 + (D(p'_i)-1)^2],
\end{align}
where $\lambda$ is a user-defined parameter to balance the losses computed on an entire point cloud and on individual points; $D(P)$ and $D(P')$ are the confidence scores returned by the discriminator $D$ when applied on generated point clouds $P$ and sampled point clouds $P'$, respectively; $p_i$ and $p'_i$ are points on $P$ and $P'$.

The generative loss $\mathcal{L}_G$ is defined as,
\begin{align}
\label{eq:SPGAN_Generative_Loss}
\mathcal{L}_G = \frac{1}{2} (D(P)-1)^2 + \beta \frac{1}{2N} \sum_{i=1}^N (D(p_i)-1)^2,
\end{align}
where $\beta$ is a user-defined parameter.

Once the generator $G$ and discriminator $D$ have been trained, the inference can be done by passing a prior latent code $z_S$ to $G$ to generate a point cloud $P=G(z_S)$. Figure~\ref{fig:dense_correspondence} illustrates generated results of the SP-GAN.

Note that the SP-GAN can not only generate point clouds but also make point-wise correspondences between the input sphere and its generated point cloud. This enables shape editing via latent code manipulation.
\section{Our Method} 
\label{sec:method}

\subsection{Overview}

Given a point cloud $P \in \mathbb{R}^{N\times 3}$ where $N$ is the number of points, we aim to learn a mapping function that maps $P$ to the latent space of the SP-GAN~\cite{li2021spgan}. Such a mapping is expected to maintain high-quality reconstruction while enabling point-wise correspondences between the input point cloud and its reconstructed point cloud.

Our GAN inversion framework consists of a point cloud encoder $E$, which learns a global latent code for an input 3D point cloud, and a generator $G$, which is the generator of the SP-GAN~\cite{li2021spgan}. 
There are three steps in the inversion as shown in Figure~\ref{fig:inversion_framework}: (1) training the encoder $E$ and the generator $G$ ($G$ can also be fine-tuned); (2) resolving point ordering; and (3) refining the global latent code extracted by the encoder $E$ into a set of local latent codes. Let us detail each step in the following sections.

\subsection{Step 1: Global latent code}

Our goal in this step is to train the encoder $E$ and the generator $G$ to learn a global latent code for each input point cloud $P$. The global latent code should be faithfully mapped to its target point cloud. This global latent code should also be invariant to each point in the target point cloud. Conceptually, the global latent code of a point cloud is similar to the global feature vector aggregated from pooling all per-point features in the point cloud as used in point cloud networks, e.g., PointNet~\cite{qi2017pointnet}. 


Suppose that the generator $G$ is given and fixed. To train the encoder $E$, we solve the following optimization problem:
\begin{equation}
\theta ^{*}_{E}= \underset{\theta _{E}}{\arg\min}\sum_{P} L\left ( G\left ( E \left ( P ; \theta _{E} \right )\right ) , P \right ),
\label{learning_based_inversion}
\end{equation}
where $L$ denotes the distance between a target point cloud and a generated point cloud. Here the generator $G$ is based on the pretrained SP-GAN generator~\cite{li2021spgan}. 

To improve the reconstruction quality, we also update the parameters of $G$ during the inversion. This can be achieved by training both $E$ and $G$ simultaneously, i.e., solving the following problem:
\begin{equation}
\theta ^{*}_{G}, \theta ^{*}_{E}= \underset{\theta_{G}, \theta _{E}}{\arg\min}\sum_{P} L\left ( G\left ( E \left ( P ; \theta_{E} \right ); \theta_{G}\right ), P \right )).
\label{learning_based_inversion_updated_G}
\end{equation}
The parameter $\theta_{G}$ and $\theta_{E}$ are updated alternatively by using the gradient descent optimization technique. 

Once the training of $E$ and $G$ is done, the global latent code $z_g$ can be determined as
\begin{align}
\label{eq:latent_code}
    z_g = E(P; \theta^*_E).
\end{align}
The global latent code $z_g$ calculated in Eq.~(\ref{eq:latent_code}) is also referred to as the prior latent code in the SP-GAN. This code is then used to generate local latent codes capturing details of the generated point cloud.

\paragraph{Encoder architecture.} There are some design choices for the architecture of the encoder $E$. Similar to image-based GAN inversion~\cite{zhu2020domain}, we can use a point cloud discriminator for the encoder. The output of the encoder is the last feature layer of the discriminator. Another option is to use a pretrained point cloud network such as DGCNN~\cite{wang2019dgcnn}, where we aggregate all local point features into a global feature vector. 
We empirically found that DGCNN yields better results, and hence choosing it for the encoder $E$. 

\paragraph{Loss functions.} Our loss function aims to enforce the learning process towards the reconstruction quality, i.e., both input and generated point clouds should have similar geometric structure, and point density. To model such similarity, we utilize the Chamfer discrepancy (CD) for our loss. In particular, we define:
\begin{equation}
\begin{aligned}
\label{eq_Chamferdistance}
L_{CD}(P, P')
=\max \bigg\{ \frac{1}{|P|} \sum_{\mathbf{p}_i \in P} \min_{\mathbf{p}'_j \in P'} \Vert \mathbf{p}_i - \mathbf{p}'_j \Vert_2,\\
\frac{1}{|P'|} \sum_{\mathbf{p}'_j \in P'} \min_{\mathbf{p}_i \in P} \Vert \mathbf{p}'_j - \mathbf{p}_i \Vert_2  \bigg\}.
\end{aligned}
\end{equation}
where $P$ is the input point cloud and $P'$ is the point cloud generated by the generator $G$. 

\begin{figure*}
\centering
\includegraphics[width=0.9\linewidth]{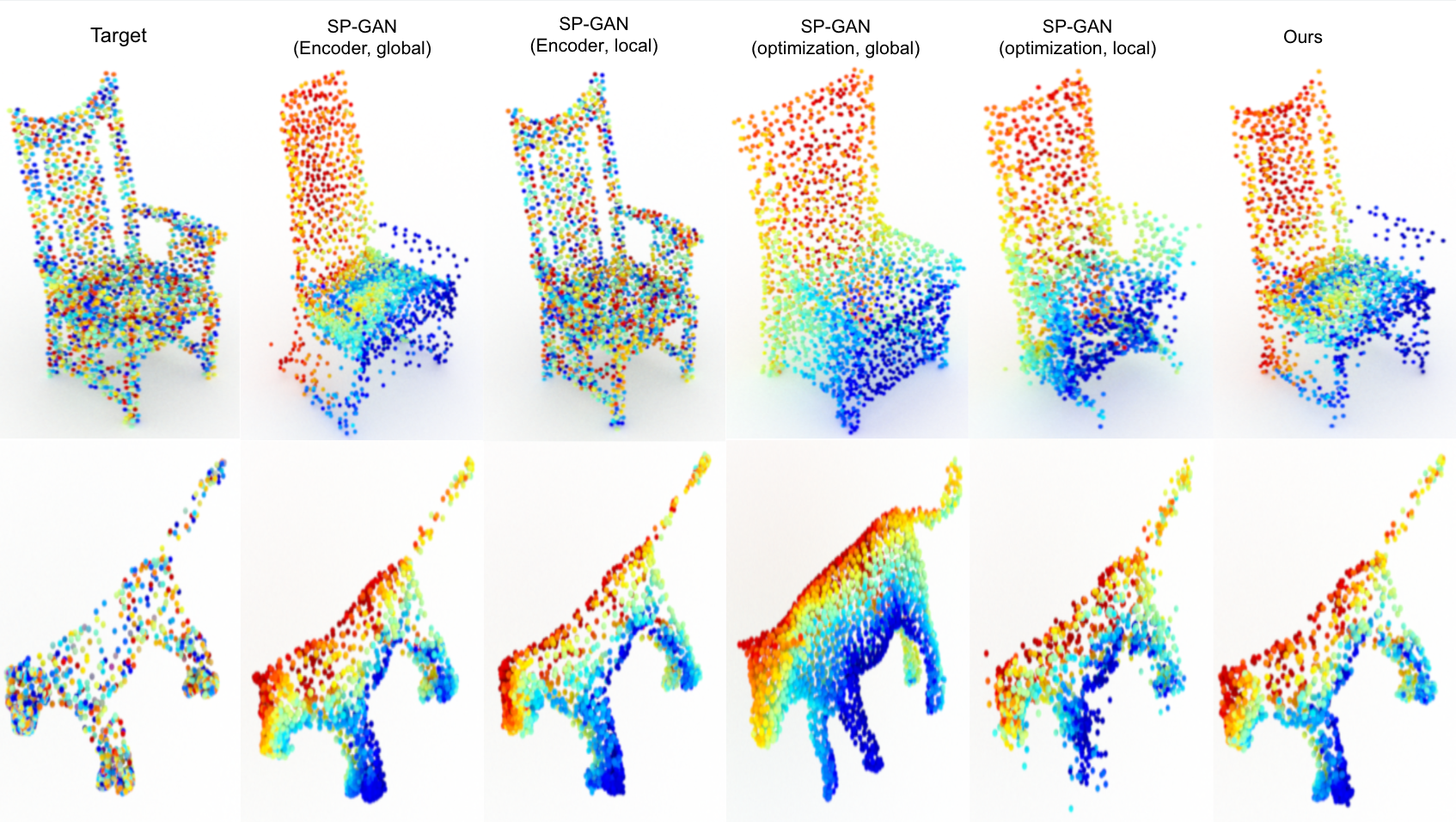}
\caption{Design choices of the latent codes. Global latent codes are invariant to point orders and thus well preserve point correspondences given in the shape prior, but produce less faithful reconstruction details. Learning-based methods using encoders to learn local latent codes tend to over-fit, resulting accurate geometry reconstruction but wrong correspondences. Optimization-based methods however incur inaccurate reconstruction. Our inversion achieves accurate geometry while preserving point correspondences.}
\label{fig:global_vs_local}
\end{figure*}
 
\begin{table*}[t]
\begin{center}
\caption{Comparison of our method with existing works in point cloud reconstruction.}
\label{table:evaluation_chamfer}
\begin{tabular}{l|c|cccc}
\toprule
 & avg. & chair & airplane & car & lamp \\
\midrule
Achlioptas et al.~\cite{achlioptas2018latent} & $3.46 \times {10^{-3}}$ & $3.61 \times {10^{-3}}$ & $1.15 \times {10^{-3}}$ & $1.14 \times {10^{-3}}$ & $7.95 \times {10^{-3}}$ \\
Zhang et al.~\cite{zhang2021shapeinversion} & $2.50 \times {10^{-3}}$ & $2.09 \times {10^{-3}}$ & $3.59 \times {10^{-3}}$& $1.95 \times {10^{-3}}$ & $2.38 \times {10^{-3}}$\\
Ours & $\mathbf{0.54 \times {10^{-3}}}$ & $\mathbf{0.66 \times {10^{-3}}}$ & $\mathbf{0.49 \times {10^{-3}}}$ & $\mathbf{0.55 \times {10^{-3}}}$ & $\mathbf{0.49 \times {10^{-3}}}$\\ 
\bottomrule
\end{tabular}
\end{center}
\end{table*}

\begin{table*}[t]
\begin{center}
\caption{Ablation studies. }
\label{table:ablation_chamfer}
\begin{tabular}{l|c|cccc||c}
\toprule
 & avg. & chair & airplane & car & lamp & animal\\
\midrule
Learning-based, global & $2.23 \times {10^{-3}}$ & $2.11 \times {10^{-3}}$ & $0.94 \times {10^{-3}}$ & $1.87 \times {10^{-3}}$ & $4.03 \times {10^{-3}}$ & $2.23 \times {10^{-3}}$\\
Learning-based, local & $0.62 \times {10^{-3}}$ & $0.59 \times {10^{-3}}$ & $0.35 \times {10^{-3}}$ & $0.62 \times {10^{-3}}$ & $0.31 \times {10^{-3}}$ & $1.27 \times {10^{-3}}$ \\
Optimization-based, global & $45.5 \times {10^{-3}}$ & $13.5 \times {10^{-3}}$ & $73.1 \times {10^{-3}}$ & $94.9 \times {10^{-3}}$ & $17.6 \times {10^{-3}}$ & $28.4 \times {10^{-3}}$ \\
Optimization-based, local & $21.2 \times {10^{-3}}$ & $2.60 \times {10^{-3}}$ & $23.4 \times {10^{-3}}$ & $48.6 \times {10^{-3}}$ & $2.77 \times {10^{-3}}$ & $7.48 \times {10^{-3}}$\\
Ours & $0.63 \times {10^{-3}}$ & $0.66 \times {10^{-3}}$ & $0.49 \times {10^{-3}}$ & $\mathbf{0.55 \times {10^{-3}}}$ & $0.49 \times {10^{-3}}$& $\mathbf{0.98 \times {10^{-3}}}$\\ 
\bottomrule 
\end{tabular}
\end{center}
\end{table*}

\begin{table}[t]
\begin{center}
\caption{Comparison of generators in reconstruction.}
\label{table:compare_generator_chamfer}
\begin{tabular}{l|cHHHH}
\toprule
 & avg. & chair & airplane & car & lamp\\
\midrule
tree-GAN~\cite{shu2019treegan} (encoder, global) & $2.64 \times {10^{-3}}$ & $2.21 \times {10^{-3}}$ & $1.29 \times {10^{-3}}$ & $2.47 \times {10^{-3}}$ & $4.61 \times {10^{-3}}$ \\
SP-GAN (encoder, global) & $2.24 \times {10^{-3}}$ & $2.11 \times {10^{-3}}$ & $0.94 \times {10^{-3}}$ & $1.87 \times {10^{-3}}$ & $4.03 \times {10^{-3}}$ \\
Ours & $\mathbf{0.54 \times {10^{-3}}}$ & $0.66 \times {10^{-3}}$ & $0.49 \times {10^{-3}}$ & $\mathbf{0.55 \times {10^{-3}}}$ & $0.49 \times {10^{-3}}$\\ 
\bottomrule 
\end{tabular}
\end{center}
\end{table}

\subsection{Step 2: Point ordering}
We found that the inversion with the global latent vector $z_g$ is limited in reconstructing geometric details of the target point cloud. Despite such, a strong advantage of using the global latent code is that it is invariant to point ordering in the target point cloud. This leads to the effect that the point-wise correspondences between the generated point cloud and the input point cloud is preserved as in the original SP-GAN~\cite{li2021spgan}.

To make the reconstruction useful for downstream tasks, it is necessary to improve the reconstruction accuracy. A naive approach is to let the encoder learn how to output local latent codes that vary for each point in the point cloud. Unfortunately, a caveat of doing so is that the ordering of the latent codes become dependent on the point ordering of the target point cloud, which could be random. This destroys the point-wise correspondences between the generated point cloud and the input shape. 
We demonstrate this issue in Figure~\ref{fig:global_vs_local}. 

Additionally, predicting the local latent codes directly might make the encoder and generator overfitted, i.e., the latent codes can just contain the target 3D coordinates and the generator simply passes these coordinates as its output. 
This yields very accurate reconstruction, but the local latent codes are useless for downstream tasks.

To improve the reconstruction quality while addressing the point ordering problem, we constrain the global and local latent codes in an engaging manner. Recall that the global latent code is produced by the encoder $E$ and is invariant to initial point ordering in a target point cloud. However, the global latent code is shown to have poor reconstruction ability. In the SP-GAN, shape prior already includes point orders, and the generator operates point-wise. However, the point orders in the shape prior can be different from those in the target shape in inversion. Therefore, to ensure consistent point ordering, we replicate the global latent code predicted by the encoder trained in Step 1 to build initial local latent codes of size $N \times (3 + d)$. This initial local latent codes are then refined in Step 3. 




\subsection{Step 3: Local latent codes}
After resolving point ordering, we are now ready to refine the global latent code into local latent codes. In this step, we keep all the parameters of the generator $G$ fixed, and update each entry in the latent code accordingly. The optimization hence becomes finding:
\begin{align}
z^{*}= \underset{z}{\arg\min}\sum_{P}^{}L\left ( G\left ( z; \theta_{G} )\right ) , P \right ),
\label{eq:optimized_based_inversion}
\end{align}
where $z$ is initialized by replicating the global latent code $z_g$.
The output of this optimization is the final local latent codes of our inversion. 


\begin{figure*}[t]
\centering
\includegraphics[width=0.9\linewidth]{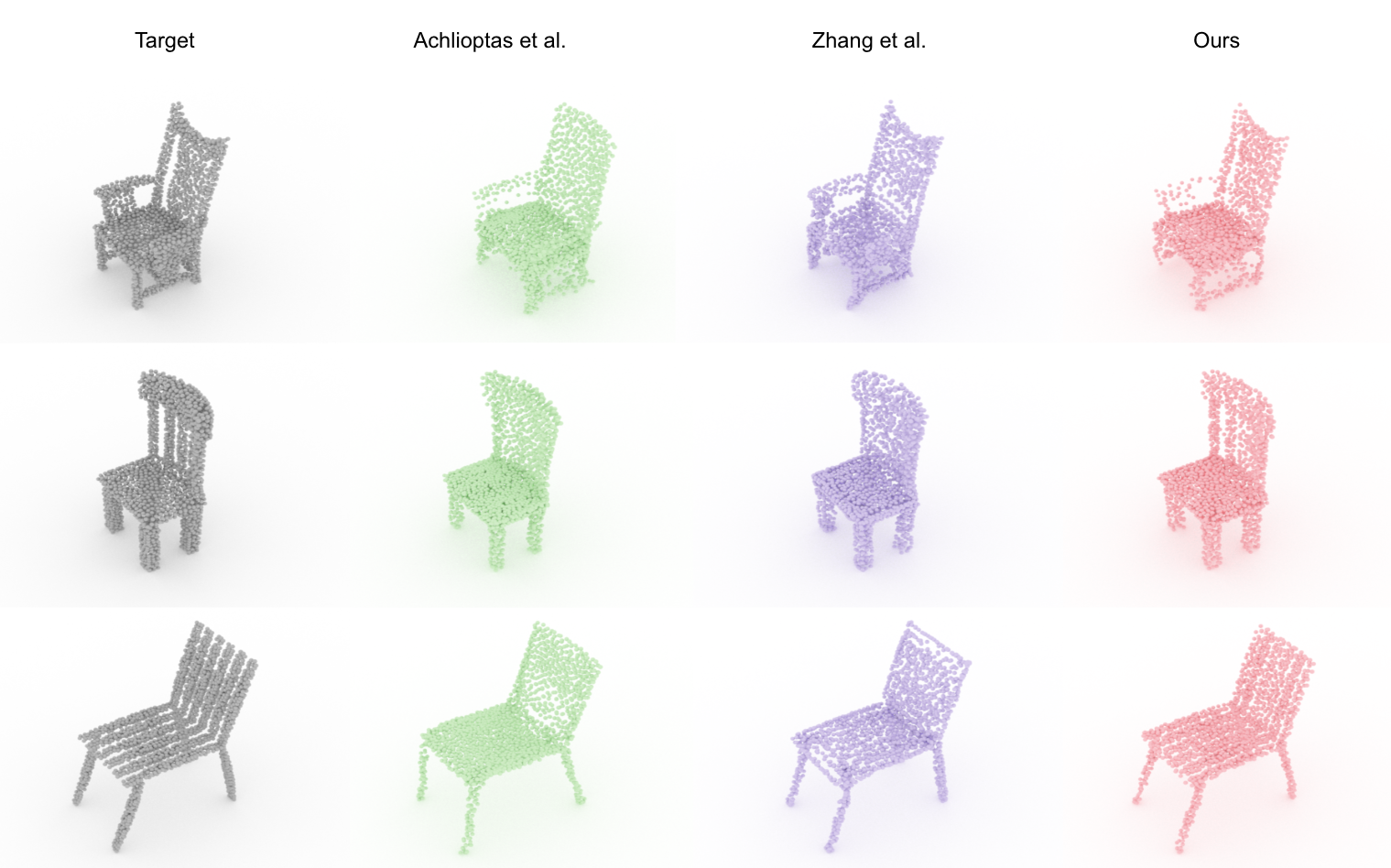}
\caption{Comparison of shape inversion examples. Our method reproduces the target more faithfully, e.g., see the back of the chairs.}
\label{fig:Visualization}
\end{figure*}

\section{Experimental Results} 
\label{sec:experiment}

\subsection{Experiment Setup}

\paragraph{Datasets.} We conducted experiments for our GAN inversion method and existing ones on the ShapeNet dataset in~\cite{chang2015shapenet}. We trained our network architecture on four object categories including chair, airplane, car, and lamp, separately. The test set was made of 10\% of the total dataset. Particularly, the chair category consists of 6,101 models for training and 677 models for testing. Beyond man-made objects in the ShapeNet, we also tested our method on the Animal dataset in~\cite{Zuff2017SMAL}. We uniformly sampled 2,048 points on object meshes to create point clouds. Like SP-GAN, we trained a single model on a combined category of all four-leg animal data such as dogs, big cats, hippos, and horses.

\paragraph{Implementation details.}
We adopted the SP-GAN in~\cite{li2021spgan} as our pre-trained generator. Our encoder was built upon the pre-trained DGCNN in~\cite{wang2019dgcnn}. We concatenated features from four layers in the DGCNN to construct the final layer of our encoder. We used 2,000 iterations for training the encoder in Step 1, and 2,000 iterations for optimizing the latent codes in Step 3 in our method. We used the Adam optimiser with a learning rate of 0.01.

\subsection{Evaluation of shape inversion}

We evaluated our method based on its reproduction quality. The reproduction quality was measured via the Chamfer discrepancy between a given target point cloud and the corresponding generated point cloud. 



\paragraph{Quantitative results.}
We first compared our method with existing shape inversion methods, e.g., the methods by Achlioptas et al.~\cite{achlioptas2018latent} and by Zhang et al.~\cite{zhang2021shapeinversion}.
The method by Achlioptas et al.~\cite{achlioptas2018latent} uses latent codes in the latent space of an autoencoder for point cloud generation. Here we adopted their autoencoder for comparison of the methods in reconstruction.
The method by Zhang et al.~\cite{zhang2021shapeinversion} is an optimization-based inversion applied to tree-GAN~\cite{shu2019treegan}. 

We report the results of this experiment in Table~\ref{table:evaluation_chamfer}. As shown in the results, our method achieves the smallest average and per-class Chamfer discrepancy. 
The reconstructed point cloud by Achlioptas et al.~\cite{achlioptas2018latent} is not editable. Compared with the results of Zhang et al.~\cite{zhang2021shapeinversion}, our reconstruction results are also more accurate by a large margin.

\paragraph{Qualitative results.}

We visually assess the quality of reconstructed point clouds by our method in  Figure~\ref{fig:Visualization}. As can be seen, our inversion algorithm can reconstruct target point clouds reasonably while preserving shape details better than other methods. For example, patterns on the back of chairs can be well recognized in our reconstructions.

\subsection{Ablation studies}
\paragraph{Global vs local latent codes.}
To further understand the effectiveness of our method, we provide an ablation study in Table~\ref{table:ablation_chamfer} and Figure~\ref{fig:global_vs_local}.
Specifically, we built different baselines including learning-based (i.e., using learned encoders) and optimization-based baselines. For each baseline, we output global or local latent codes, which are then used by the SP-GAN for point cloud reconstruction. 
Table~\ref{table:ablation_chamfer} shows that the learning-based baselines are generally better than the optimization-based ones. Additionally, optimizing local latent codes increases the precision of reconstruction.
However, it is worth noting that this could lead to overfitting, as shown in the case of using encoders to output local codes. In this case, reconstruction achieves the best accuracy but point correspondences are significantly corrupted in reconstructed shapes (see Figure~\ref{fig:global_vs_local}), making subsequent shape manipulation impossible. Our method instead has slightly lower reconstruction accuracy compared with the overfitting case, but it can keep the correspondences intact. We found that dense correspondence problem is not shown in the animal dataset. This is probably because the number of static shapes in the animal dataset is small while the shapes less diverge. The encoder can avoid overfitting, but reconstruction results are not as good as those from the ShapeNet. Note that our method guarantees the reconstruction performance and dense correspondence regardless of the datasets.

Table~\ref{table:ablation_chamfer} also shows the results on the Animal dataset. As can be seen, our method outperforms all the baselines. Moreover, the results of all the methods on this dataset match the performance trend reported on the ShapeNet. A visualization of the results on the Animal dataset is in Figure~\ref{fig:global_vs_local}.

\begin{table}[t]
\begin{center}
\caption{Comparison of encoders in inversion on the chair class.}
\label{table:encoder_model_comparison}
\begin{tabular}{c|cc}
\toprule
DGCNN~\cite{wang2019dgcnn} & SP-GAN's discriminator~\cite{li2021spgan}\\
\midrule
$5.43 \times {10^{-4}}$ & $6.92 \times {10^{-4}}$\\
\bottomrule 
\end{tabular}
\end{center}
\end{table}

\paragraph{Generator.} We also experimented with our GAN inversion with different generators. We chose tree-GAN~\cite{shu2019treegan} and compared it with SP-GAN~\cite{li2021spgan}. Comparison results are shown in Table~\ref{table:compare_generator_chamfer}. As can be seen, SP-GAN remains more effective than tree-GAN in reconstruction, even taking a simple setting including an encoder and a global latent code. SP-GAN also has better point correspondences compared to tree-GAN. 

\paragraph{Encoder architecture.}
We evaluated the design choice of our encoder by comparing the inversion accuracy of two architectures: DGCNN~\cite{wang2019dgcnn} and the discriminator in the SP-GAN~\cite{li2021spgan}. Note that the discriminator in the SP-GAN also makes use of convolutional layers for feature learning and thus can be used as an encoder. Results of this experiment are shown in Table~\ref{table:encoder_model_comparison}. 
It can be seen that the DGCNN has a smaller Chamfer discrepancy than the SP-GAN's discriminator. Moreover, we empirically observed that the use of DGCNN makes the training of the encoder converge much faster. This consolidates our choice of the DGCNN for the encoder in our architecture.

\begin{figure}[t]
\centering
\includegraphics[width=0.8\linewidth]{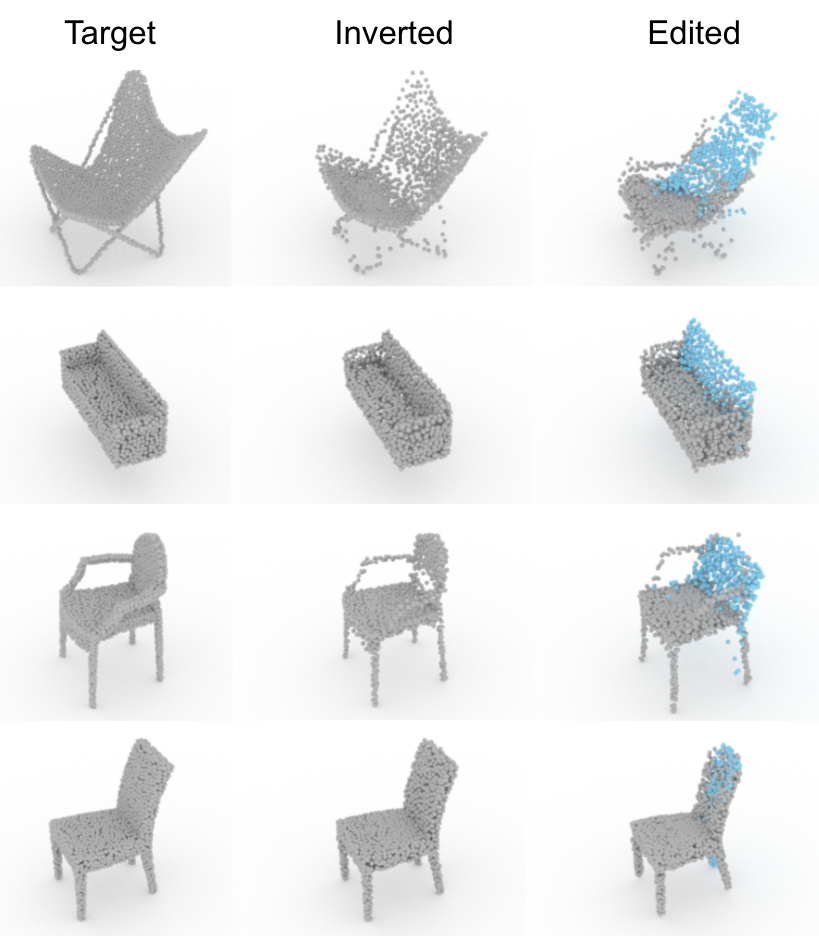}
\caption{Shape editing of the back of the chairs. Our method allows the change of the back of a chair from a rectangular to a round shape and vice versa. The colored part is the updated geometry.}
\label{fig:edit_chair_back}
\end{figure}

\begin{figure}[t]
\centering
\includegraphics[width=0.8\linewidth]{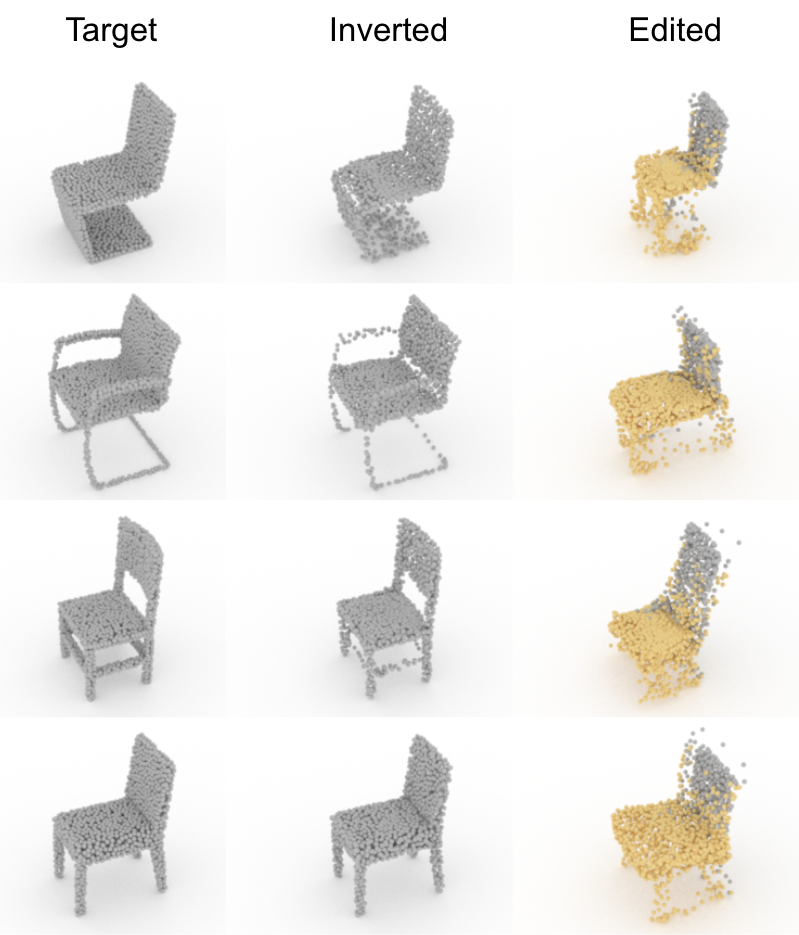}
\caption{Shape editing of the seats and legs of the chairs. Note the changes in sizes and styles compared to the target point clouds. The colored part is the updated geometry.}
\label{fig:edit_chair_seat}
\end{figure}

\subsection{Application: Shape editing}
SP-GAN~\cite{li2021spgan} is suitable for shape editing because it learns the dense correspondence implicitly. Therefore, after inversion, each (semantic) part of the generated point cloud should correspond to a region in the shape prior. 
As our method can maintain point-wise correspondences, we can utilize such correspondences to enable \emph{part-aware} shape manipulation.
Note that previous methods, e.g.,~\cite{zhang2021shapeinversion}, only demonstrate shape manipulation by global jittering of the latent codes. Here, we segment regions of interest in the point cloud that we would like to manipulate, then perturb the corresponding local latent codes of points in those regions to obtain new local latent codes. The final shape can be generated by passing the perturbed codes to the generator. We demonstrate this capability in Figure~\ref{fig:edit_chair_back} and Figure~\ref{fig:edit_chair_seat}.

As shown in Figure~\ref{fig:edit_chair_back}, we can alter the style of the back of chairs, e.g., changing a chair's back from a rectangular shape to a round shape, or making a sofa's back longer.
In Figure~\ref{fig:edit_chair_seat}, we showcase the changes in chair seat size and chair leg style.

Our shape editing is not perfect. We observed that the local latent codes might be entangled, i.e.,  changing a part might lead to incidental changes in other parts. Disentangling the shape latent space is left for our future work.

\vspace{0.3cm}
\section{Conclusion} 
\label{sec:conclusion}

We proposed a new point cloud GAN inversion method that allows faithful reconstruction of 3D point clouds using a sphere-guided point cloud generator~\cite{li2021spgan} while maintaining point correspondences during inversion. Our method outperforms existing GAN inversion works in terms of reconstruction quality, verified both quantitatively and qualitatively. We demonstrate the usefulness of our inversion method via a shape editing task, i.e., editing reconstructed point clouds by manipulating part-aware latent codes. 

Our work is not without limitations. First, our reconstruction might still miss some small details in  target point clouds. Further research in improving the reconstruction quality is thus worthwhile. Second, the latent code of the SP-GAN is not compact. Exploring GAN inversions with compact latent space would benefit a wider range of downstream applications. Finally, it is worth applying the latent codes for more downstream applications such as shape completion from real scans~\cite{chen2020pcl2pcl}. Extending the inversion to colored point clouds would also be an interesting research avenue.

\noindent\textbf{Acknowledgment.} This paper was partially supported by an internal grant from HKUST (R9429).

{\small
\bibliographystyle{ieee_fullname}
\bibliography{egbib,ganinversion}
}

\clearpage
\appendix
\renewcommand*\appendixpagename{\Large Supplementary Materials}
\appendixpage
\begin{abstract}
In this supplementary material, we provide visualizations of the reconstructed and edited point clouds of more categories (airplane, car, lamp, animal) (Figure 1-8). We also show the evaluation of our reconstruction quality in the EMD metric (Table~\ref{table:evaluation_emd}).
\end{abstract}

\section{Additional Visual Results}
In this section, we visualized more categories such as airplane, car, lamp and animal to compare our proposed method with existing methods. We also use the reconstructed latent codes to manipulate point clouds. As can be seen, our method can reconstruct objects faithfully while predicting smooth correspondences. Note that there is no correspondence in original target shapes to the shape prior of the generator, demonstrated by the noisy color rendition. And Fig~\ref{fig:treeGAN_shapenet_Visualization}, ~\ref{fig:treeGAN_animal_Visualization} show the reconstruction visualization of treeGAN~\cite{shu2019treegan} based global encoder baseline. The inversion performance is lower than SP-GAN~\cite{li2021spgan} based inversion model.

\section{Additional Evaluation on EMD}
We evaluate the reconstruction quality of our method and existing ones using the Earth Mover's Distance (EMD). As can be seen in Table~\ref{table:evaluation_emd}, overall, our proposed method has the smallest distances to the ground truth, especially in airplane and car.

\begin{figure*}[t]
\centering
\includegraphics[width=1.0\linewidth]{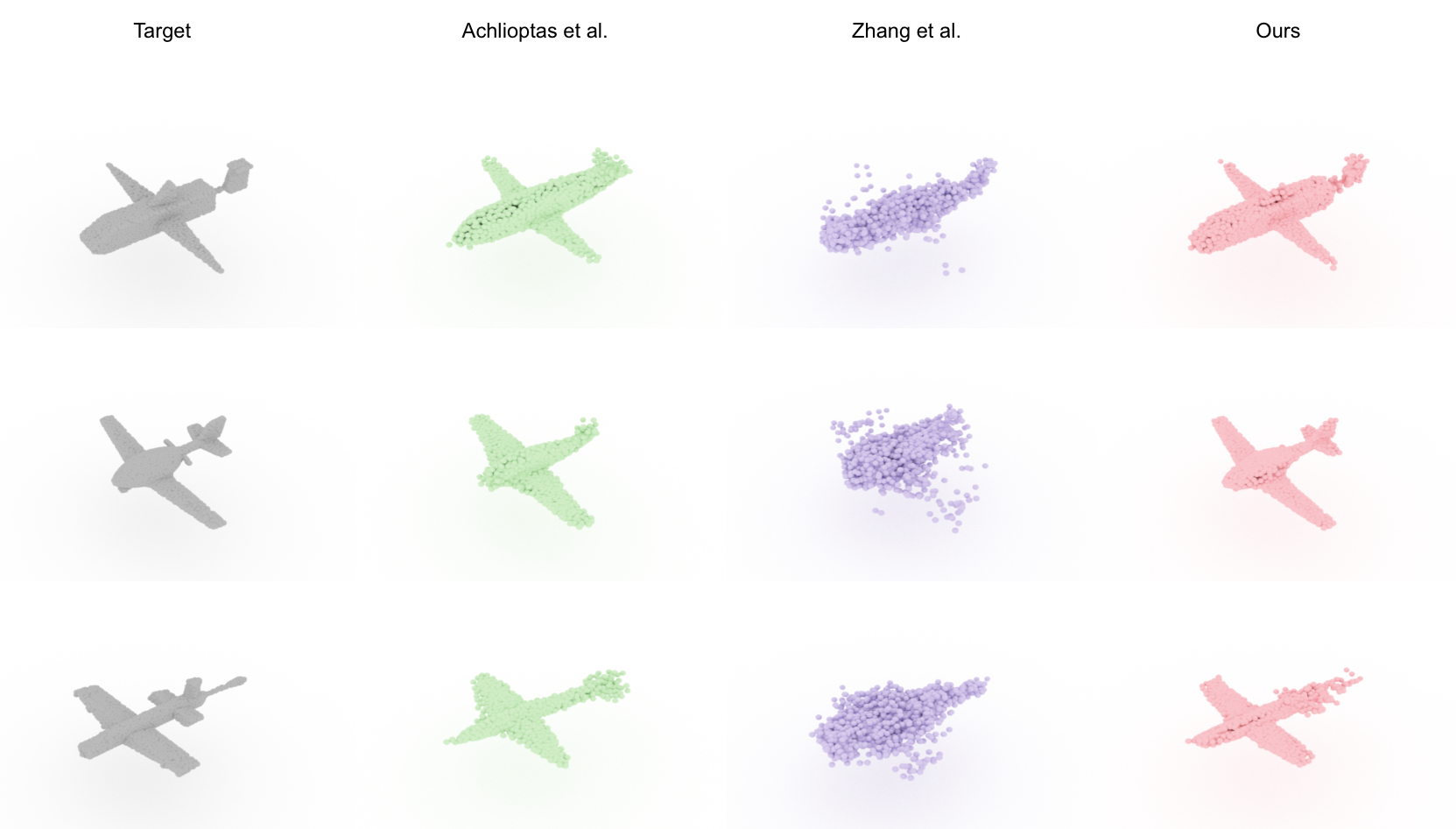}
\vspace{0.05in}
\caption{Inversion examples generated by our method and existing methods (airplane). Our method reproduces the target more faithfully.}
\vspace{0.2in}
\label{fig:airplane_Visualization}
\end{figure*}


\begin{figure*}[t]
\centering
\includegraphics[width=1.0\linewidth]{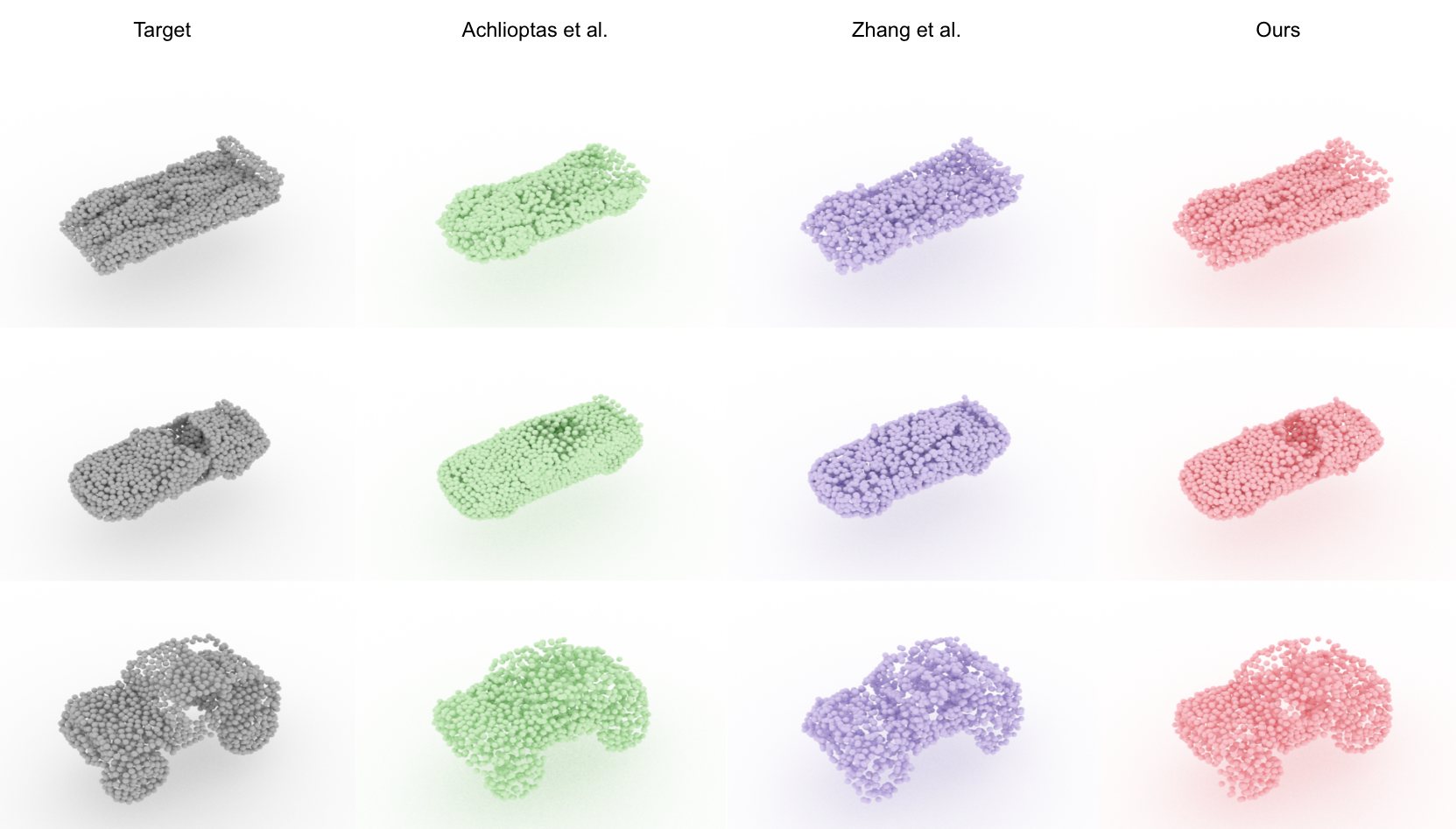}
\vspace{0.05in}
\caption{Inversion examples generated by our method and existing methods (car). Our method reproduces the target more faithfully.}
\vspace{0.2in}
\label{fig:car_Visualization}
\end{figure*}


\begin{figure*}[t]
\centering
\includegraphics[width=1.0\linewidth]{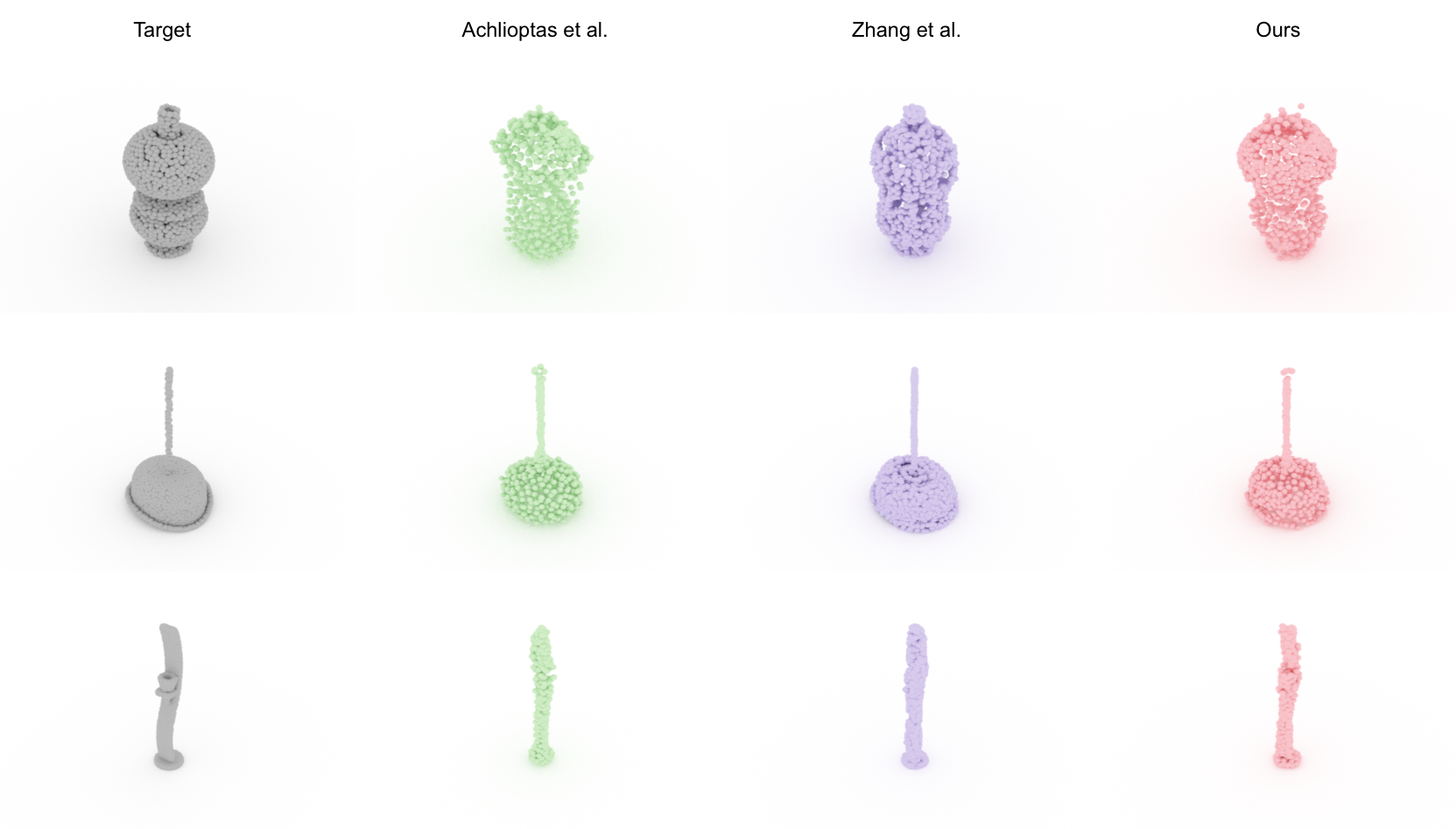}
\vspace{0.05in}
\caption{Inversion examples generated by our method and existing methods (lamp). Our method reproduces the target more faithfully.}
\vspace{0.2in}
\label{fig:lamp_Visualization}
\end{figure*}
\begin{figure*}[t]
\centering
\includegraphics[width=1.0\linewidth]{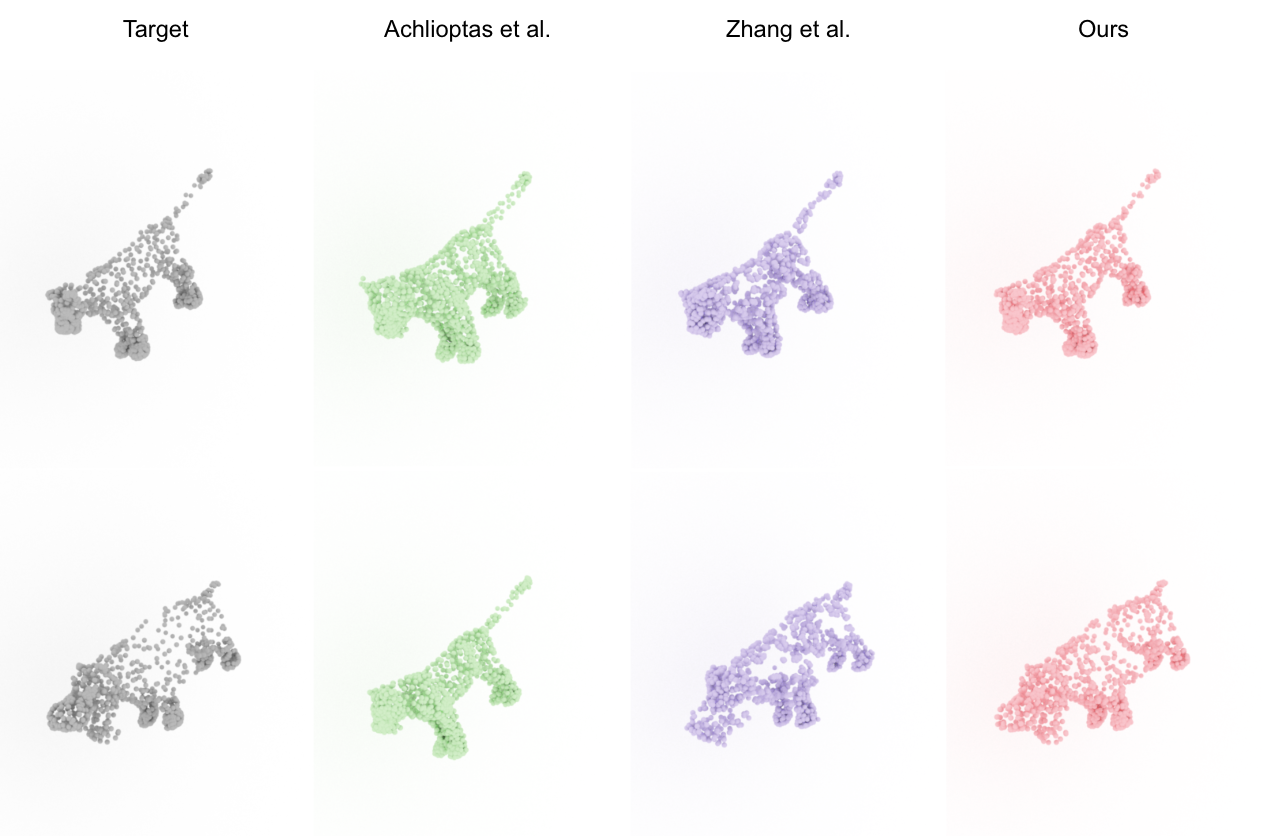}
\vspace{0.05in}
\caption{Inversion examples generated by our method and existing methods (animal). Our method reproduces the target more faithfully.}
\vspace{0.2in}
\label{fig:animal_Visualization}
\end{figure*}

\begin{figure*}[t]
\centering
\includegraphics[width=0.9\linewidth]{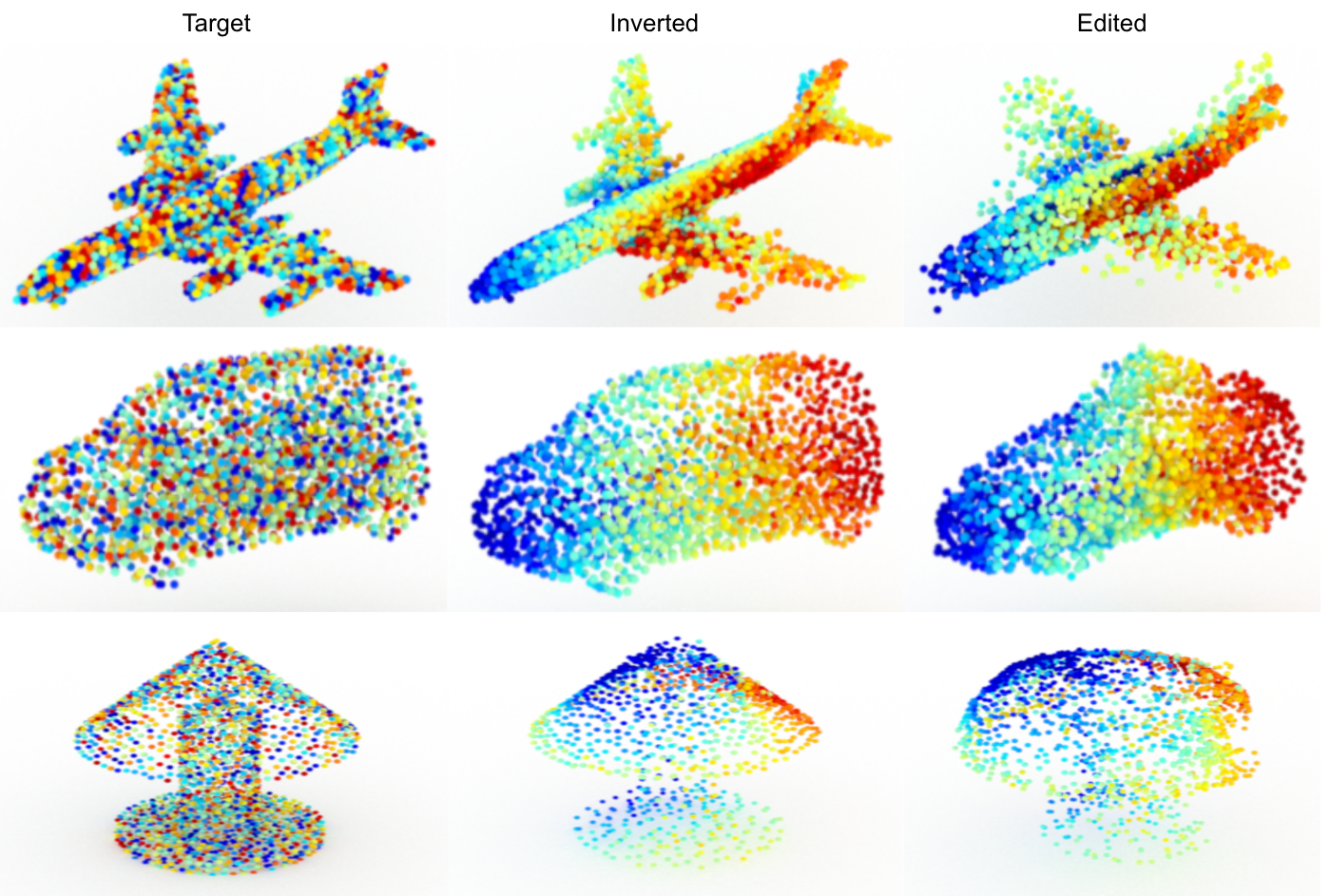}
\vspace{0.05in}
\caption{Inversion and edited point clouds (Airplane, Car, Lamp)}
\vspace{0.2in}
\label{fig:edited_airplane}
\end{figure*}
\begin{figure*}[t]
\centering
\includegraphics[width=0.9\linewidth]{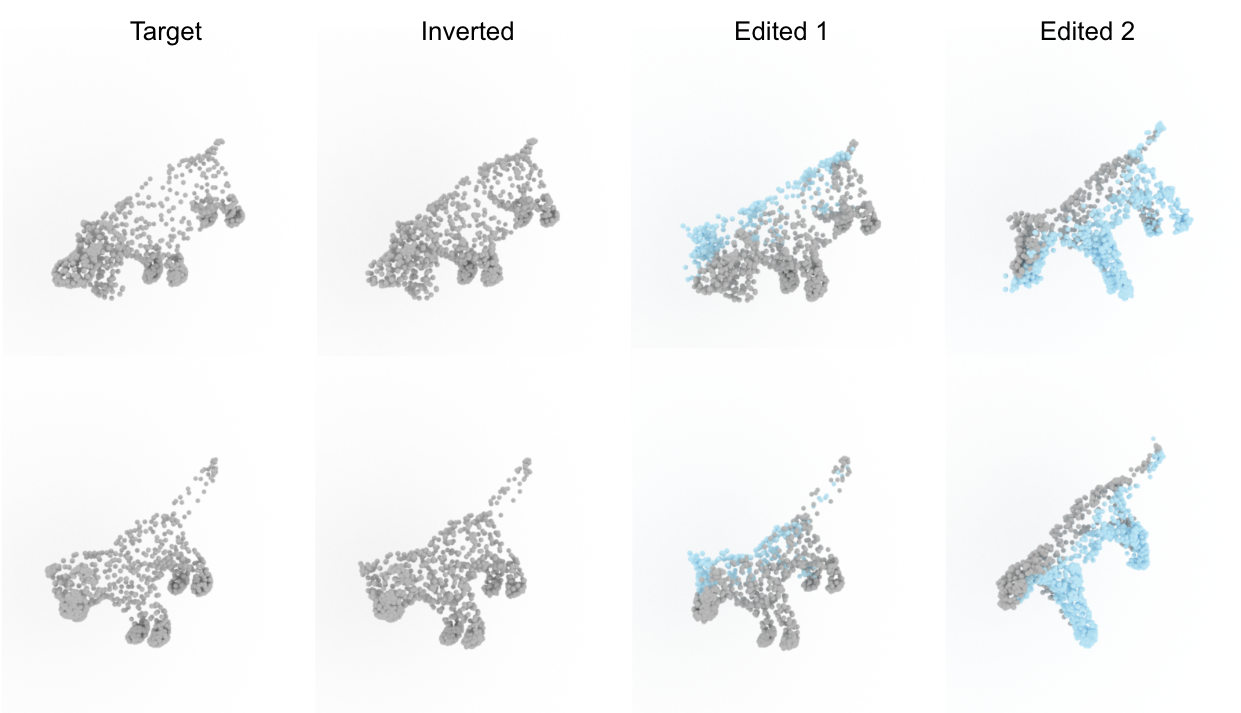}
\vspace{0.05in}
\caption{Inversion and edited point clouds (Animal)}
\vspace{0.2in}
\label{fig:edited_airplane}
\end{figure*}

\begin{figure}[t]
\centering
\includegraphics[width=1.0\linewidth]{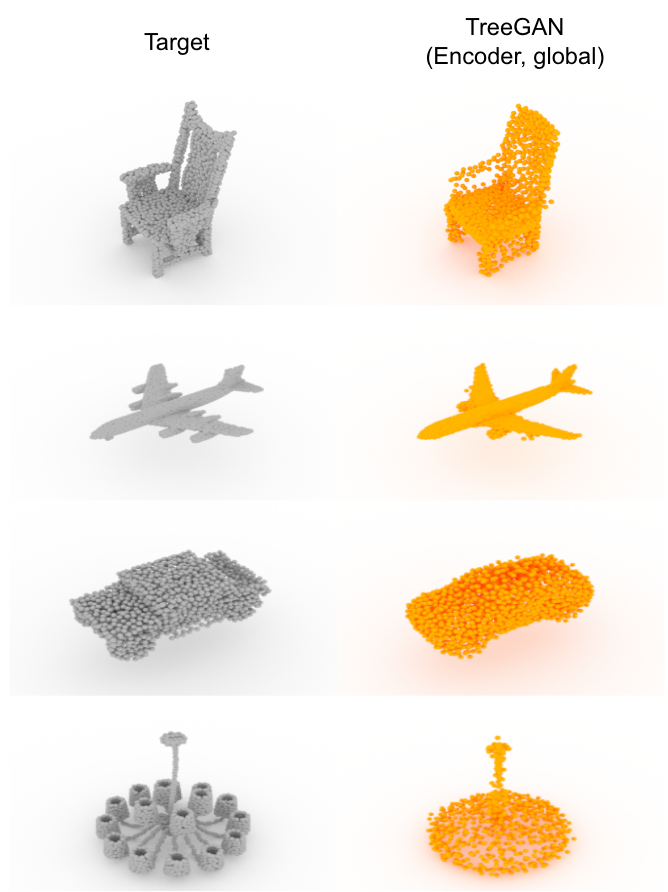}
\vspace{0.05in}
\caption{Inversion examples generated by treeGAN global encoder baseline.}
\vspace{0.2in}
\label{fig:treeGAN_shapenet_Visualization}
\end{figure}

\begin{figure}[t]
\centering
\includegraphics[width=1.0\linewidth]{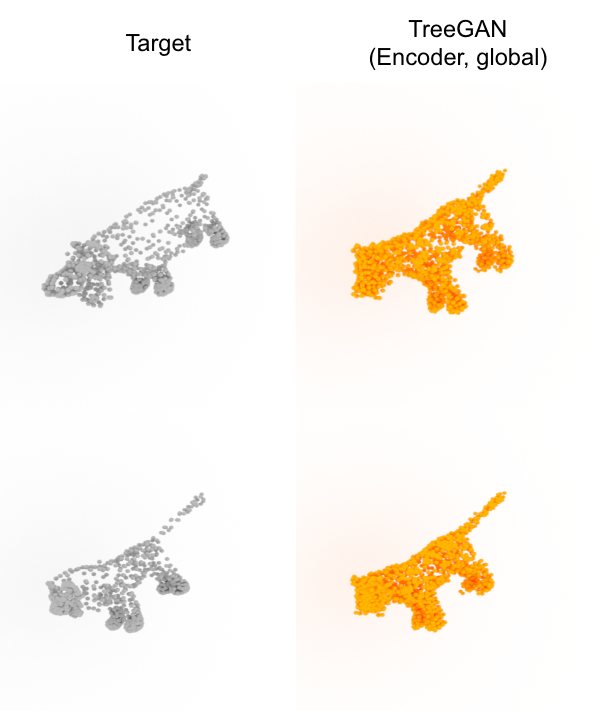}
\vspace{0.05in}
\caption{Inversion examples generated by treeGAN global encoder baseline.}
\vspace{0.2in}
\label{fig:treeGAN_animal_Visualization}
\end{figure}

\begin{table*}[t]
\begin{center}
\caption{Evaluation of the reconstruction quality using EMD metric.}
\label{table:evaluation_emd}
\vspace{0.3cm}
\begin{tabular}{l|c|c|c|c|c|c}
\toprule
 & chair & airplane & car & lamp & animal\\
\midrule
Base & $3.82 \times {10^{-2}}$ & $2.86 \times {10^{-2}}$ & $3.77 \times {10^{-2}}$ & $8.99 \times {10^{-2}}$& $5.61 \times {10^{-2}}$\\
Zhang et al.~\cite{zhang2021shapeinversion} & $4.46 \times {10^{-2}}$ & $6.15 \times {10^{-2}}$ & $3.72 \times {10^{-2}}$& $6.71 \times {10^{-2}}$& $5.81 \times {10^{-2}}$\\
Ours & $4.90 \times {10^{-2}}$ & $2.19 \times {10^{-2}}$ & $2.44 \times {10^{-2}}$ & $8.06 \times {10^{-2}}$ & $3.61 \times {10^{-2}}$\\
\bottomrule 
\end{tabular}
\end{center}
\end{table*}

\end{document}